\pdfoutput=1

\documentclass[11pt]{article}
\usepackage{authblk}

\usepackage{ifthen}
\newboolean{acl}

\setboolean{acl}{false}
	
\ifthenelse{\boolean{acl}}{\usepackage[review]{EMNLP2023}}{\usepackage{EMNLP2023}}

\usepackage{times}
\usepackage{latexsym}
\usepackage{soul}
\usepackage{xspace}

\usepackage{caption}
\usepackage{subcaption}

\usepackage[T1]{fontenc}

\usepackage[utf8]{inputenc}

\usepackage{microtype}

\usepackage{inconsolata}

\usepackage{graphicx}
\usepackage{amsmath}
\usepackage{tabularx}
\usepackage{multicol}
\usepackage{multirow}
\usepackage{algorithm}
\usepackage{algpseudocode}
\usepackage{inconsolata}
\usepackage{tikz}
\usepackage{tcolorbox}
\usepackage{enumitem}
\usepackage{booktabs}
\usepackage{subfiles}
\usepackage{xcolor}
\usepackage{pbox}

\newcommand\blfootnote[1]{%
	\begingroup
	\renewcommand\thefootnote{}\footnote{#1}%
	\addtocounter{footnote}{-1}%
	\endgroup
}

\newcommand{\redhl}[1]{\setlength{\fboxsep}{0.25pt}\colorbox{pink}{#1}}
\newcommand{\greenhl}[1]{\setlength{\fboxsep}{0.25pt}\colorbox{green!20}{#1}}

\usetikzlibrary{shapes.geometric, arrows}

\tikzstyle{startstop} = [rectangle, rounded corners, minimum width=3cm, minimum height=1cm,text centered, draw=black, fill=orange!30]
\tikzstyle{io} = [trapezium, trapezium left angle=70, trapezium right angle=110, minimum width=3cm, minimum height=1cm, text centered, draw=black, fill=blue!30]
\tikzstyle{process} = [rectangle, minimum width=3cm, minimum height=1cm, text centered, draw=black, fill=green!30]
\tikzstyle{decision} = [diamond, aspect=3, minimum width=3cm, minimum height=1cm, text centered, draw=black, fill=red!30]
\tikzstyle{arrow} = [thick,->,>=stealth]
\usepackage[framemethod=tikz]{mdframed}

\usepackage{xspace,mfirstuc,tabulary}

\usepackage{microtype}
\newcounter{notecounter}

\newcommand{\enoteson}{\long\gdef\enote##1##2{{
			\stepcounter{notecounter}
			{\large\textbf{ \hspace{1cm}\arabic{notecounter} $<<<$ ##1: ##2 $>>>$\hspace{1cm}}}}}}
\enoteson

\def\ModelName{LongForm\xspace}

\def\DataName{LongForm-C\xspace}
\def\MethodNames{Reverse Instructions\xspace}
\def\methodnames{reverse instructions\xspace}
 
\def\methodname{reverse instructions\xspace}

\title{\ModelName:
Effective Instruction Tuning with \MethodNames}

\author[*$\diamond$$\dag$]{Abdullatif Köksal}
\author[*]{Timo Schick}
\author[$\dag$]{Anna Korhonen}
\author[*$\diamond$]{Hinrich Sch\"utze}

\affil[*]{Center for Information and Language Processing, LMU Munich}
\affil[$\diamond$]{Munich Center for Machine Learning}
\affil[$\dag$]{Language Technology Lab, University of Cambridge \protect\\
\texttt{akoksal@cis.lmu.de}}

\long\def\devour#1{\ignorespaces}

\mdfdefinestyle{MyFrame}{%
    linecolor=black,
        outerlinewidth=2pt,
                innertopmargin=4pt,
                    innerbottommargin=4pt,
                        innerrightmargin=4pt,
                            innerleftmargin=4pt,
                                    leftmargin = 4pt,
                                            rightmargin =
        4pt
                    }

\begin{document}
	\maketitle \begin{abstract} Instruction tuning enables language models to
		more effectively generalize and better follow user
		intent. However, obtaining instruction data is costly and
		challenging. Prior work employs methods such as expensive
		human annotation, crowd-sourced datasets with alignment
		issues, and generating noisy examples via LLMs. We introduce
		the \DataName dataset, which is created by \textit{\methodnames}.  
		We generate instructions via LLMs for human-written corpus examples using \methodnames. 
		First we select a
		diverse set of human-written documents from corpora such as
		C4 and Wikipedia; then we generate instructions for these
		documents via LLMs. This approach provides a cheaper and
		cleaner instruction-tuning dataset with natural output and one suitable for long
		text generation. 
		Our models
		outperform 10x larger language models without instruction
		tuning on tasks such as story/recipe generation and
		long-form question answering. Moreover, \ModelName models
		outperform prior instruction-tuned models such as FLAN-T5
		and Alpaca by a large margin, and improve language understanding
		capabilities further. 
		We publicly release
		our data and models: \ifthenelse{\boolean{acl}}
		{[Anonymized-URL]}
		{\url{https://github.com/akoksal/LongForm}}.
	  \end{abstract}

	 \ifthenelse{\boolean{acl}}{}{\blfootnote{First published on ArXiv on 2023-04-17. This version extends the training with recent LLMs, evaluation with new metrics, and NLU tasks.}}

	\begin{figure}[!t]
		\centering
		\includegraphics[width=0.9\linewidth]{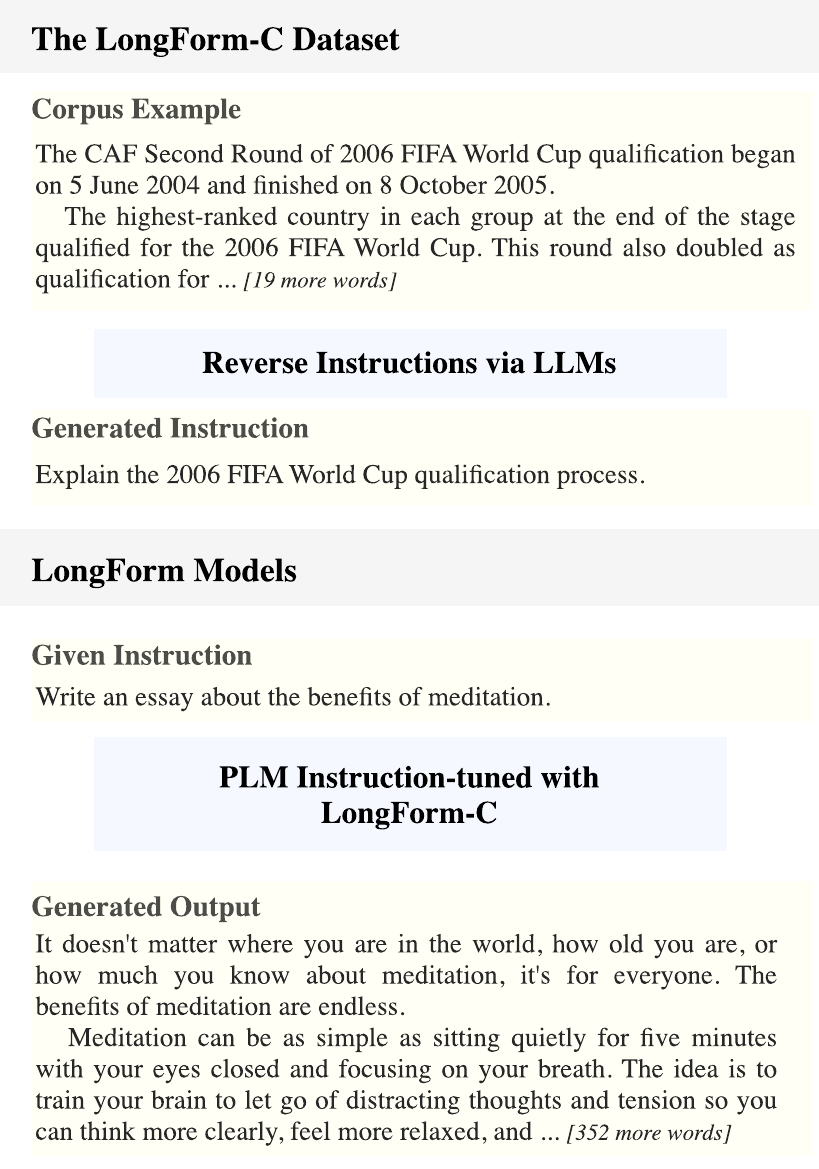}
		\caption{
Our \textbf{\MethodNames} method
extracts high-quality human-written passages from
		corpora (``Corpus Example'') and then
		leverages an LLM to generate, for each
		corpus example, a \emph{generated instruction}
		that should elicit this corpus example.
                Using our method, we create
the 
			\DataName instruction tuning
		dataset,
                consisting of 
                \textit{(generated instruction, corpus
		example)}
                 pairs. We then instruction-tune a PLM
with \DataName. The PLM output shown (``Generated Output'')
indicates that this approach produces
		high-quality output.
Evaluated on a diverse set of text generation
		tasks,  \MethodNames outperforms 
			prior models by more than 35\%.}

		\label{fig:intro_example}
		\end{figure}

\devour{
		\caption{Our \textbf{\MethodNames} method creates the 
			\DataName dataset: LLMs generate instructions for given high-quality human-written
			texts.
			\ModelName models are
			instruction-tuned over these \textit{(generated instruction, corpus example)} pairs and
			evaluated on a diverse set of text generation tasks, outperforming 
			prior models by more than 35\%.}
}

	\section{Introduction}
	Instruction tuning conditions language models (LMs)
        on
user intents and
        improves cross-task generalization
while ensuring better human alignment. Recent
        works on instruction tuning focus on data
        collection, showing that training with even a small amount of
        instruction-tuning data can outperform much larger
        LMs in following user intent and
        generalization \cite{ouyang2022training,
        chung2022scaling}. However, these studies have some
        limitations, such as reliance on expensive
        human-annotated instruction
        data \cite{ouyang2022training} or
focusing on
instructions for academic NLP tasks that
have limited coverage of
real-world generation tasks \cite{chung2022scaling, wang-etal-2022-super,
        sanh2022multitask}. A third approach, works that
        generate instructions and outputs from scratch using
        large language models (LLMs), often produces
        low-quality
        datasets \cite{honovich2022unnatural, wang-etal-2022-super}.

Using the \textbf{\methodnames} method,
we create an instruction-following text generation dataset
called \textbf{\DataName} to address these issues.
We aim to improve
instruction tuning by gathering diverse human-written texts
from C4 and English Wikipedia as outputs. As illustrated in
Figure \ref{fig:intro_example}, we extract
text passages
(paragraphs or
documents) from the corpora and
then prompt an LLM
with a zero-shot
template to generate, for each passage, an instruction that -- when given to an
LLM -- we want to result in the generation of the passage
or a similar text. We refer to this method of creating
\textit{(generated instruction, passage)} pairs
for instruction tuning
as \textbf{\methodnames}.
This method offers
a cost-effective and fast alternative for creating
instruction tuning datasets compared to human annotation
while also yielding higher quality output compared to
fully synthetic data generation.

To make \textbf{\DataName} even more effective,
we increase
its diversity and quality by
leveraging structured examples from
Stack Exchange and Wi\-ki\-How, and long text generation tasks
from NLP benchmarks.

Based on
the \DataName dataset,
we finetune
	instruction-following PLMs, which we
	call \textbf{\ModelName} models, with different
	architectures and sizes:
	T5-XL \cite{JMLR:v21:20-074_raffel_t5}, OPT-6.7B \cite{zhang2022opt}, and
	LLaMA-7B \cite{touvron2023llama}. We
	compare these models with the baselines
	 FLAN-T5 \cite{chung2022scaling},
	T0pp \cite{sanh2022multitask},
	T\textit{k}-Instruct \cite{wang-etal-2022-super},
	and Alpaca on a diverse set of tasks: story, poem,
	email and recipe generation; grammar error
	correction; text summarization;
table-to-text;
and long form question
answering. Our experimental results are as follows.
(i) \ModelName-OPT-2.7B
outperforms OPT-30B on long text generation 
despite OPT-30B having 10x more parameters. (ii)
\ModelName models outperform
prior instruction-following models, such as FLAN-T5
and Alpaca, with more than \textbf{63\%} (in-domain)
and \textbf{35\%} (out-of-domain) relative METEOR improvement
on text generation. (iii)
In addition to NLG,
the \DataName{} dataset also enhances
the language understanding (NLU) capabilities of
several LLMs. 
It improves performance on MMLU \cite{hendryckstest2021} compared 
to prior datasets like FLAN.

(iv) 
For multilingual news generation, \ModelName models follow
	multilingual instructions and generate news articles
	in German,
                Spanish, French and Russian better than
	prior models.

We release  the \DataName{}
	dataset and
        models publicly on \ifthenelse{\boolean{acl}}
        {[Anonymized-URL]}
        {\url{https://github.com/akoksal/LongForm}}.

	\section{Related Work}
	\textbf{Instruction-following models}: The
	instruction paradigm was introduced to better
	control language models using natural language
	commands \cite{ouyang2022training,
	wang-etal-2022-super, chen-etal-2022-mtg,
	wei2022finetuned}. While some
	approaches \cite{ouyang2022training} focus on
	obtaining costly human-annotation data through their
	platform  and improving human alignment with
	reinforcement learning, recent
	works \cite{srivastava2022imitation,
	wang-etal-2022-super} have extended this paradigm to
	various existing NLP tasks by reformulating
	them. This approach helps to overcome the
	instruction-data bottleneck, but is limited to
	academic tasks. Additionally, this approach is still
	restricted in terms of text generation: only 7 out
	of 1762 tasks in
	BigBench \cite{srivastava2022imitation} and 12 out
	of 1613 in Super-Natural Instructions
	(NIv2) \cite{wang-etal-2022-super} include English
	tasks with long outputs (i.e., average number of
	words
	$>$50). Self-Instruct \cite{wang2022selfinstruct}
	and Unnatural
	Instructions \cite{honovich2022unnatural} challenge
	academic benchmarks by generating instruction data
	(input, instruction and output) via LLMs to
	provide a more diverse dataset. However, relying
	solely on generation via LLMs
    has drawbacks: the overall validity of Self-Instruct is only 54\% \cite{wang2022selfinstruct}. More recent works focus on curating costly human-annotated datasets such as Dolly \cite{DatabricksBlog2023DollyV2}.

	\noindent\textbf{Data Generation with LMs}: Recent
	neural network based approaches to NLP rely on large
	and diverse training datasets. Thus, many works
	focus on augmenting or generating new training
	examples.
        For example, existing datasets are expanded  to
	enhance model quality for question answering (QA) \cite{longpre-etal-2019-exploration}, part-of-speech tagging \cite{sahin-steedman-2018-data}, and textual similarity \cite{schick-schutze-2021-generating}. For a more general-purpose training, Self-Instruct \cite{wang2022selfinstruct} and Unnatural Instructions \cite{honovich2022unnatural} propose the use of LLMs to generate tasks and examples including instructions and outputs. WizardLM \cite{xu2024wizardlm} focuses on evolving human-written instructions by prompting LLMs and generating more complicated instructions that require multiple-step reasoning.
	 To the best of our knowledge, \DataName is the first work to combine corpora and LLMs to automatically generate a general-purpose text generation dataset.
	
	\noindent\textbf{Corpus Mining}: Most NLP tasks
	benefit from extracting useful examples from corpora
	such as machine translation (MT) with bitext
	mining \cite{resnik-1999-mining} and argument
	mining. Recent approaches use embeddings
	to find similar texts in different languages to mine
	human translation from unlabeled
	data \cite{artetxe-schwenk-2019-margin}. These
	methods have been extended to MT with up to 200
	languages \cite{nllbteam2022language}. Additionally,
	some tasks filter sentences from corpora to create
	diverse training sets for human annotation, such as
	argument mining \cite{ein2020corpus}. Recent works
	explore methods to mine and restructure corpora
	(e.g., Wikipedia)   to generate synthetic data for
	conversational QA \cite{zhunyun2022dialog} and
	closed-book
    QA \cite{lewis-etal-2021-paq}. 
	
	\noindent\textbf{Long Text Generation}: Long text
	generation is challenging. It requires
	models to understand long dependencies and planning
	to generate long texts. Many works do
	task-specific generation
(e.g.,  generation of
	stories \cite{fan-etal-2018-hierarchical},
	recipes  \cite{bien-etal-2020-recipenlg},
	essays \cite{feng2018topic} and long-form QA \cite{fan-etal-2019-eli5})
 by collecting useful examples from existing resources
	(e.g., ELI5 subreddit, WritingPrompts subreddit, recipe websites).
 Also, proper evaluation of long text generation remains a significant challenge for task-specific or general-purpose models \cite{celikyilmaz2020evaluation}.

\ifthenelse{\boolean{acl}}{}{
 \noindent\textbf{Reverse Instructions}:
 The \methodnames{}
 method offers a cost-effective and faster alternative for creating instruction 
 tuning datasets compared to human annotation, while also
 yielding higher quality output compared to
 fully-synthetic data generation. Building upon this approach,
 \citet{wang2023harnessing} and \citet{li2023selfalignment} expand
 the concept of \methodnames{} by integrating instruction data filtering mechanisms 
 and open-source models. Furthermore, \citet{chen2023tegit} 
 introduces a rewriting step of corpus samples in \methodnames{}.
 In a significant extension to multilingual settings, \citet{koksal2024muri} proposed MURI (Multilingual Reverse Instructions), which leverages multilingual corpora and translation models to generate instruction-following datasets for 200 languages. Moreover, \citet{ziegler2024craft} introduced CRAFT, which extends the \methodnames{} methodology to task-specific contexts through the generation of structured datasets via corpus example retrieval and LLM-based reformulation. 
 }
 
 \noindent The \methodnames{} methodology constitutes an efficacious and cost-efficient alternative to conventional human annotation for instruction tuning dataset creation, while demonstrating superior quality compared to fully-synthetic data generation approaches. This framework has been subsequently extended through several significant contributions in the literature. Specifically, \citet{wang2023harnessing} and \citet{li2023selfalignment} augmented the fundamental \methodnames{} paradigm through the integration of instruction data filtering mechanisms and the utilization of open-source models. \citet{chen2023tegit} further advanced this methodology by introducing a corpus sample rewriting procedure within the \methodnames{} framework. \citet{koksal2024muri} expanded it to multilingual settings via MURI (Multilingual Reverse Instructions), which leverages multilingual corpora with translation models and LLMs to generate instruction-following datasets for 200 languages. Furthermore, \citet{ziegler2024craft} developed CRAFT, adapting the \methodnames{} approach to generate task-specific structured datasets through corpus example retrieval and reformulation with LLMs.

	\begin{figure*} \centering \includegraphics[width=0.98\linewidth]{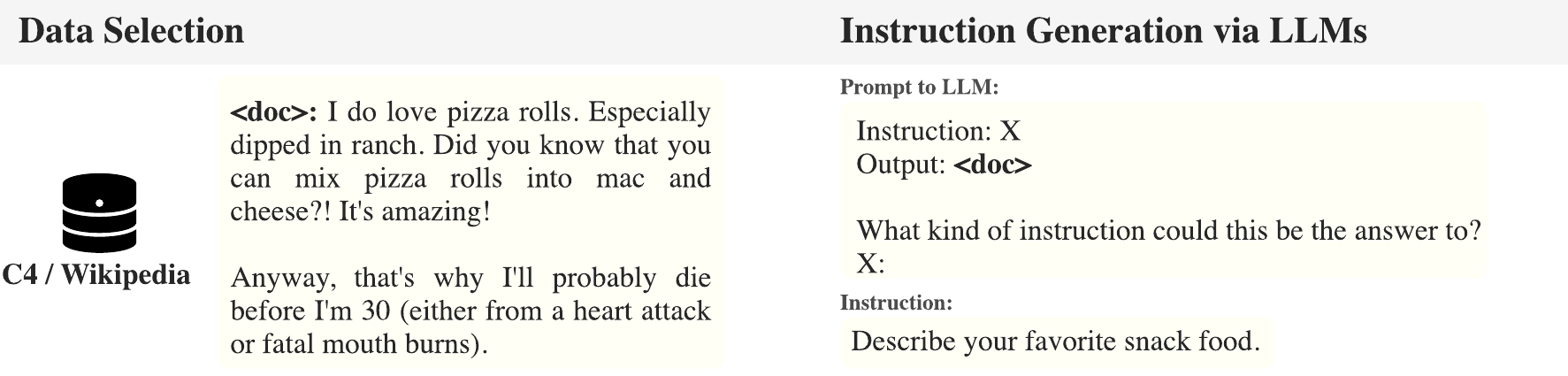} 
		\caption{
			The \methodnames method. After collecting diverse examples
			from corpora, we generate relevant
			instructions through zero-shot prompting via LLMs in
			various styles. (see \S\ref{sec:llm_query_appendix})}  \label{fig:dataset_example} \end{figure*}
	
	\section{The \DataName{} Dataset}
	\label{sec:dataset}
The \DataName{} dataset  consists of 15,000
corpus examples generated via \methodnames{} and additional
	12,739 examples from structured corpora (SC) 
and NLP datasets with instructions and long text pairs.
The main part of the
\DataName{} dataset is constructed by \methodname{}  generation for a diverse set
of corpus samples; this part is referred to as \textbf{RI} below. Additionally, we expand our dataset with
structured corpora examples through parsing and templates
(referred to as \textbf{SC}), as
well as NLP tasks reformulated to increase diversity
(referred to as \textbf{NLP}). Thus,
our dataset features a diverse collection of instructions
(including human-written, template-based and LLM-generated), each
paired with a human-written output.

	\subsection{\MethodNames (RI)}

        Figure \ref{fig:dataset_example} summarizes the \methodnames method.

1. \textbf{Data Selection}: We sample 10,000 examples from
the C4 corpus and 5,000 examples from the English Wikipedia
for selecting target text in English. Since C4 is
noisy \cite{dodge-etal-2021-documenting}, we choose only
those texts whose URLs had received three or more upvotes in
Reddit, following \citet{radford2019language}. However, this
filtering may result in a lack of diversity, as most URLs
may originate from specific subreddits that are more active,
such as the news subreddit. %
To address this, we k-means cluster
all documents in C4 based on their BERT
embeddings. We then select 10,000 examples, one example from
each cluster that is closest to the cluster center. This
helps  prevent corpus examples from being dominated by
multiple samples from a specific domain or cluster, thus
ensuring a greater diversity of texts.

For the English Wikipedia, we adopt a direct approach since
it is already diverse. We randomly select 5,000
articles and extract the first paragraph for 75\%
 and the first two paragraphs for 25\% of
the examples. The resulting dataset contains
shorter texts:  average length is 57$\pm$43 words 
(compared to 408$\pm$263  for C4). 
The overall average in \DataName{} is 291$\pm$272.
	
	2. \textbf{Instruction Generation}: We aim to
	generate relevant instructions for the 15K corpus
	examples using LLMs. We employ GPT3
	(text-davinci-003 from OpenAI API) as our LLM. We
	design prompts for generating instructions for a
	given document and query the LLM in a zero-shot
	manner. We create three templates to diversify the
	styles of our instructions: formal instruction style
	(50\%), informal chatbot style (30\%), and search
	engine query style (20\%). The formal instruction template is structured as follows:  
	\begin{mdframed}[style=MyFrame,hidealllines=true,backgroundcolor=gray!5]
		\small
		Instruction: X  
		
		\noindent Output: ``<corpus\_example>''
		
		\noindent What kind of instruction could this be the answer to?  
		
		\noindent X:
	\end{mdframed}
        \vspace{-0.2cm}
	We add ``X:'' at the end of the template as a prompt
        to generate plausible instructions.
See \S\ref{sec:llm_query_appendix} for
details of LLM parameters and templates.

Finally, we
incorporate length information into the generated
instruction using a predefined set of templates to
signal the desired length of the output and provide
additional control to the model. As the length of
our corpus examples varies (291$\pm$272 words), we
provide templates such as ``Respond in D sentences''
or ``Respond in D words''.
We
also include less precise templates such as
``Respond briefly''
(resp.\ ``Respond in detail'')
for outputs of less than 3 (resp.\ more than 10)
sentences.
We append or prepend these
templates to the original instructions for
a 30\%
subset of our dataset.
This lets us control \ModelName models in terms of length.

\subsection{Structured Corpora (SC)}
We also mine
structured corpora that contain (instruction,(long-)output)
pairs: Stack Exchange and WikiHow.
We do not use LLMs here.
	
From the
	\textbf{Stack Exchange (SE)}
	subcorpus of Pile \cite{gao2020pile}, we select 50
	examples from each of 88 subdomains, covering a wide
	range of topics, including Buddhism, chemistry, and
	webmasters. For each example, we select the question
	and its corresponding details as an instruction and
	the answer as the output. This subset adds more
	complicated human instructions to \DataName.
	
\textbf{WikiHow} includes how-to tutorials,
        each comprising a
	question and an answer that consists of two parts: an
	introduction and several steps with a brief summary and
	a long description of each step.
\devour{
We initially used
the WikiHow summarization
	dataset \cite{koupaee2018wikihow}; however, as this
	dataset lacks details (e.g., the distinction
	between brief summary and a long description), 
	we scrape the website with more granularity.}

To generate an instruction, we create 18 templates -- e.g.,
``What are some steps to <question>?'' --
and use ``how-to'' questions to fill in the placeholders.
We also
include the number of steps in 14 out of 18 templates to
control the model in terms of number of steps, similar to the
length information for corpus-extracted examples. We provide all
templates in \S\ref{sec:wikihow_templates_appendix}.

We generate target texts by combining various
elements, including introductory paragraphs (i.e., part 1), summaries
 and long descriptions of each step (i.e., part 2). We only
 include the first part in half of the examples. %
The second part is always included, but in different formats: either as 
a summary, a long description, or both.

	\subsection{NLP Tasks (NLP)}
	 Most of instruction tuning benchmarks of NLP
	tasks involve classification or short text
	generation tasks, accounting for over 99\% of
	BigBench and NIv2. However, we opt for NLP
	tasks with long outputs to enrich \DataName. From
	BigBench, we select Helpful Answer Generation
	(hhh) \cite{askell_2021_hhh_dataset} and Minute
	Mysteries QA, a question answering dataset for short
	stories. From NIv2, we include a selection of ten
	tasks including poem
	generation \cite{hipson-mohammad-2020-poki}, table
	to  text
	generation \cite{iyyer-etal-2017-search}, story
	generation \cite{sap-etal-2020-recollection,
	orbach-goldberg-2020-facts2story,
	lin-etal-2019-reasoning}, 
	summarization \cite{fabbri-etal-2019-multi,
	kornilova-eidelman-2019-billsum,
	brazinskas-etal-2020-shot}, and fact
	generation \cite{kotonya-toni-2020-explainable-automated}. For 
a more uniform distribution across datasets,
	we sample a similar number from
	each dataset, resulting in 600 samples from BigBench
	and 3,684  from NIv2.

	Finally, we include the Enron dataset to
	create an email writing task from the subject using
	templates similar to
	WikiHow. We also employ the BEA-2019 for
	grammatical error
	correction \cite{bryant-etal-2019-bea}; 
	incorrect inputs and template instruction are the
	input, the correct output the target.
See \S\ref{sec:enron_templates_appendix} for
templates for Enron  and
	\S\ref{sec:bea_gec_templates_appendix} 
for BEA-2019.

\begin{table}[t]
	\small
	\setlength\tabcolsep{4.6pt}
	\centering
	\begin{tabular}{llr}
		\toprule
		\textbf{Type} & \textbf{Source} &\textbf{\# of Examples} \\
		\midrule
		\multirow{2}{*}{\shortstack[l]{Reverse Instructions\\(RI)}} & C4 & 10,000\\
		& Wikipedia & 5,000\\
		\cmidrule{2-3}
		\multirow{2}{*}{\shortstack[l]{Structured\\Corpora (SC)}} & Stack Exchange & 4,380\\
		& WikiHow & 2,500 \\
		\cmidrule{2-3}
		\multirow{4}{*}{NLP Tasks (NLP)} &NIv2 & 3,684 \\
		& Big Bench & 600\\
		& BEA-GEC & 1,203\\
		& Enron & 372\\
		\midrule
		\multicolumn{2}{l}{\textbf{Total}} & 27,739\\
		\bottomrule
	\end{tabular}
	\caption{Origin statistics for \DataName.
		Examples are generated with \methodname{}, from
		structured corpora (SC) and NLP
		tasks (NLP).}
	\label{tab:data_writinginstructions}
\end{table}

\begin{figure}[t]
	\centering
	\includegraphics[width=\linewidth]{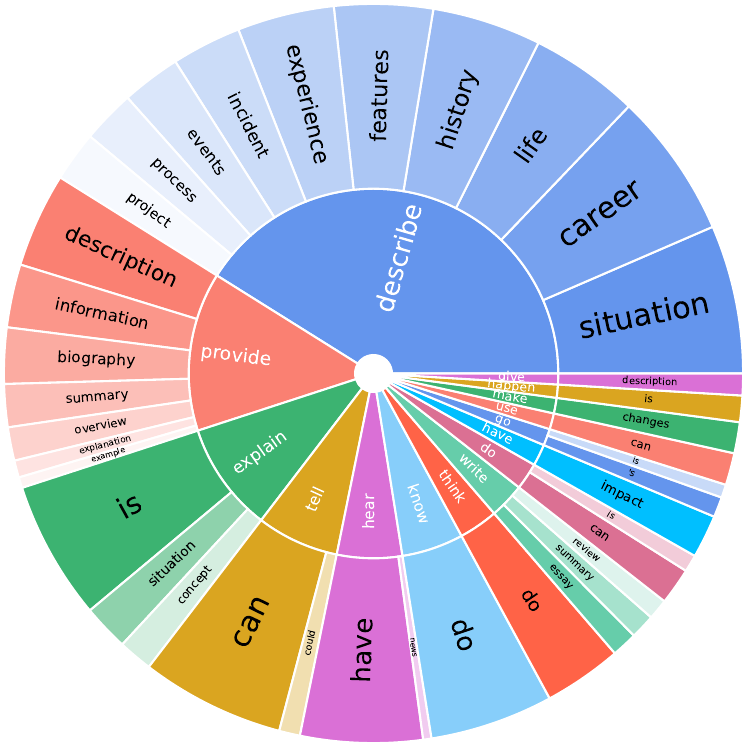}
	\caption{Most common noun+verb and
		auxiliary+verb pairs of generated
		instructions from the \methodname{} (RI) subset of \DataName. 
	}
	\label{fig:instruction_analysis}
\end{figure}

\subsection{Dataset Analysis}
	\textbf{Diversity:}
Following \citet{wang2022selfinstruct}, we
analyze the diversity of the generated
	instructions by parsing each
	instruction in the corpus subset with Berkeley
	Neural
	Parser \cite{kitaev-klein-2018-constituency}. We
	extract either the verb and its direct object
	(noun+verb: ``describe history'') or
        the auxiliary with its dependent verb
	(auxiliary+verb: ``do know'').
 8,872 of 15,000
	instructions follow this format.
	Figure \ref{fig:instruction_analysis}
visualizes the 3,167
	most common noun+verb/auxiliary+verb pairs.
They cover a diverse range of topics,
	including events, careers, biographies, and writing
	tasks such as reviews, summaries and essays -- and
	this is just looking at the 11\% of
our \DataName dataset with \methodnames (RI). For a detailed 
comparison of \DataName's instruction diversity with other datasets 
such as Dolly \citep{DatabricksBlog2023DollyV2} and FLAN \citep{chung2022scaling}, 
see Appendix (\S\ref{sec:diversity_comparison}).

        Table \ref{tab:data_writinginstructions}
shows the distribution of \DataName:
54\% of the examples are generated with \methodnames,  25\%  from
structured corpora, and
 21\% are reformulated from NLP tasks. The dataset is split
23,652/2,042/2,045 into
training/validation/test.

	\section{Experimental Setup}
	\subsection{\ModelName Models}

We create three \ModelName models by finetuning language
models on the \DataName\ dataset to assess their
	effectiveness on instruction-following text
	generation:
\textbf{\ModelName-T5-3B}, based
	on T5-3B \cite{JMLR:v21:20-074_raffel_t5,
	lester-etal-2021-power}, an
encoder-decoder LM, and two autoregressive LMs:
\textbf{\ModelName-OPT-6.7B}, based on
OPT-6.7B \cite{zhang2022opt},
and \textbf{\ModelName-LLaMA-7B}, based on 
	LLaMA-7B \cite{touvron2023llama}. We share
	the details of finetuning in the
	Appendix (\S\ref{sec:training_details}).

	\subsection{Baselines}
	We compare our instruction-tuned LMs with other LMs
	that are instruction-tuned on other datasets. These
	LMs are based on either
	T5-11B \cite{JMLR:v21:20-074_raffel_t5,
	lester-etal-2021-power} or
	LLaMA-7B \cite{touvron2023llama}. We
	also compare our LMs with a raw large LM, OPT-30B \cite{zhang2022opt}. Notably, all LMs have the same or higher number of parameters than our \ModelName models.

	\textbf{OPT} \cite{zhang2022opt} is an open-source decoder-only transformer LM, trained on diverse English corpora. Due to hardware limitations that make long text inference from LLMs time-consuming, we use the 30B variant of OPT.
	
	\textbf{T0++} \cite{sanh2022multitask} is an
	LM-adapted variant of T5-11B,
	instruction-tuned on
	PromptSource \cite{bach-etal-2022-promptsource}, a dataset of $>$12 million examples from existing datasets.
	
	\textbf{T\textit{k}-Instruct} \cite{wang-etal-2022-super}
	is another instruction-tuned LM based on an
	LM-adapted version of T5-11B. The training data is a subsample of NIv2, which includes 757 tasks and over 75,000 examples. As we are performing zero-shot long text generation via instruction, we use the definition-only version of T\textit{k}-Instruct.

	\textbf{Flan-T5} \cite{chung2022scaling} is an
        instruction-tuned LM built on the LM-adapted version
        of T5-11B. The training data
(total size 14M)
includes tasks and prompts from multiple sources, including Muffin \cite{wei2022finetuned}, T0-SF \cite{sanh2022multitask}, and NIv2 \cite{wang-etal-2022-super}.
	
	\textbf{Alpaca} is an instruction-tuned LM  built on
	LLaMA-7B \cite{touvron2023llama} by using a
	variation of
	Self-Instruct \cite{wang2022selfinstruct}. As they
	did not release their finetuned LM, we finetune
	LLaMA-7B with their variation of Self-Instruct.%
                                \footnote{\url{github.com/tatsu-lab/stanford_alpaca/}}

	\begin{table*}[t]
		\footnotesize
		\setlength\tabcolsep{4.6pt}
		\centering
		\begin{tabular}{lrrrrrrrrrr}
			\toprule
			& &    & \multicolumn{2}{c}{\textbf{RI}} &  \multicolumn{2}{c}{\textbf{SC}} &  \multicolumn{4}{c}{\textbf{NLP}} \\
			& \textbf{Model Size} &   \textbf{Avg.} & \textbf{C4} &  \textbf{Wiki} &  \textbf{SE} &  \textbf{WikiHow} & \textbf{Enron} &  \textbf{BEA-GEC} &  \textbf{NIv2} &  \textbf{BigBench} \\
		\cmidrule(lr){1-3} \cmidrule(lr){4-5} \cmidrule(lr){6-7} \cmidrule(lr){8-11}
			T0++   & 11B  &   5.9 &   2.8 &   8.1 &    2.3 &    2.8 &    4.3 &     17.1 &                  11.3 &       1.7 \\
			T\textit{k}-Instruct  & 11B  &   6.0 &   2.2 &   9.4 &    3.2 &    1.3 &    3.8 &     12.4 &                  13.5 &       2.5 \\
			Flan-T5 &  11B & 12.5 &   4.1 &  13.3 &    5.8 &    4.1 &    8.4 &     70.8 &                  17.2 &       3.2 \\
			Alpaca-LLaMA-7B & 7B & 15.2 &   8.4 &  22.2 &    9.2 &   10.7 &    6.4 &     55.3 &                  16.9 &       8.6 \\
		\cmidrule(lr){1-3} \cmidrule(lr){4-5} \cmidrule(lr){6-7} \cmidrule(lr){8-11}
			OPT-30B & 30B  &  12.3 &  14.2 &  13.7 &   14.4 &    8.0 &    7.0 &     16.8 &                   9.5 &       3.3 \\
		\cmidrule(lr){1-3} \cmidrule(lr){4-5} \cmidrule(lr){6-7} \cmidrule(lr){8-11}
			\ModelName-T5-XL & 3B  &  24.3 &  18.9 &  22.4 &   15.3 &   21.5 &    8.7 &     \textbf{92.9} &                  \textbf{22.2} &      17.8 \\
			\ModelName-OPT-6.7B & 6.7B  &  24.2 &  \textbf{19.7} &  25.0 &   15.9 &   22.0 &    \textbf{8.9} &     85.1 &                  20.8 &      15.8 \\
			\ModelName-LLaMA-7B & 7B & \textbf{24.8} &  19.6 &  \textbf{27.8} &   \textbf{16.3} &   \textbf{22.6} &    6.6 &     86.2 &                  21.2 &      \textbf{18.0} \\
			\bottomrule
		\end{tabular}
		\caption{METEOR scores of baselines and our
			LMs on each subset 
			of \DataName test
and micro
			average (``Avg.'') over  the entire test set. All \ModelName models
			outperform prior LMs on long text
			generation with a big
			margin. \ModelName-T5-XL provides
			comparable results on text
			generation tasks (BEA-GEC, NIv2, BigBench), but the best results are achieved with \ModelName-LLaMA-7B.}
		\label{tab:res_writinginstructions}
	\end{table*}

	\subsection{Evaluation}
	We evaluate the performance of the baselines and our
	instruction-tuned LMs on different sizes and
	architectures. For generation, we perform nucleus sampling \cite{Holtzman2020The} with $p=0.9$ for all LMs. We first evaluate them on the
	test set of \DataName; recall that it consists of a diverse set of NLP tasks and corpus examples.
	
	Additionally, we assess the generalization
	capabilities of our LMs on a set of long text
	generation datasets that were not seen during
	finetuning. These tasks include recipe
	generation \cite{bien-etal-2020-recipenlg} from a
	given set of ingredients, story generation for a
	given prompt from Writing Prompts (WP) subreddit,
	and long form question answering,
	ELI5 \cite{fan-etal-2019-eli5}. We sample 250
	examples from each task, resulting in 750 examples
	in total. Although these datasets are not part
	of \DataName, recipe generation is part of the
	instruction tuning data of
        two baselines, T\textit{k}-Instruct and Flan-T5.
	
	Next we analyze  long-form generation
	 and language understanding abilities. We compare the performance 
	 of LMs on out-of-domain long 
	 text generation tasks (i.e., RGen, ELI5, WP) and MMLU \cite{hendryckstest2021}.
	 We compare LMs finetuned with the \DataName{} dataset and
	 the FLAN collection,\footnote{\citet{wang2023far}'s
	 FLAN version  \cite{longpre2023flancollection}}
	 which has a strong performance on MMLU \cite{chung2022scaling}.
	 We also extend this analysis on more recent and powerful LLMs. In addition
	 to LLaMA-7B, we compare FLAN and \DataName datasets on
	  LLaMA 2-7B \citep{touvron2023llama2}
	 and Mistral 7B \citep{jiang2023mistral}.
	
	We also compare on multilingual
	text generation.
	We reformulate the multilingual
	summarization corpus
	MLSUM \cite{scialom-etal-2020-mlsum} by
	providing an instruction in French, German,
	Spanish,  Russian to generate news articles in
	those languages for given titles. We utilize the
	translation of the following instruction: ``Write a
	news article for the given title: X'' where X is the
	title and the output is the full news article. We
	collect 100 samples for each language and then
	evaluate on news article generation. We report results in
	\S\ref{sec:appendix_multilingual} in Appendix.
	
	As current metrics in text generation have limited
	capabilities in evaluating long text
	generation \cite{celikyilmaz2020evaluation}, we
	choose METEOR \cite{banerjee-lavie-2005-meteor} as
	our main metric as it exhibits higher human
	correlation \cite{sharma2017relevance,
	chen-etal-2022-mtg}. However, we also report
	BLEU \cite{papineni-etal-2002-bleu},
	ROUGE \cite{lin-2004-rouge}, and
	self-BLEU \cite{self-bleu} scores in the Appendix (\S\ref{sec:appendix_results}), which exhibit similar patterns.

	\section{Results}
We now present a comprehensive evaluation of our
\ModelName models compared to the baselines on \DataName test, recipe generation, long form question answering, short story generation, NLU, and ablation of \DataName{} subsets.
	\subsection{\DataName Results}
	We begin by comparing baseline LMs and \ModelName
	models on the test set of \DataName. As presented in
	Table \ref{tab:res_writinginstructions},
	all \ModelName models demonstrate clear improvements
	in performance compared to the prior
	instruction-tuned and raw LMs. In
	particular, \ModelName-T5-XL, with 3B parameters,
	outperforms all instruction-tuned LMs based on
	T5 (i.e., T0++, T\textit{k}-Instruct, Flan-T5), even with 11B parameters, across all subtasks. This suggests the effectiveness of \DataName in finetuning LMs to follow instructions for long text generation.
	
	Moreover, among our LMs, we see
	that \ModelName-LLaMA-7B
	outperforms \ModelName-OPT-6.7B, despite both having
	similar number of parameters, which aligns with
	findings
	from \citet{touvron2023llama}. Additionally, we
	observe that \ModelName-T5-XL has comparable or
	better results than auto-regressive variants,
	particularly on text generation  (i.e., BEA-GEC,
	NIv2, BigBench). However, a potential limitation of T5 variants
	is the absence of newline tokens in their vocabulary,
	which may impact the readability of the generated
	output, particularly for long text generation.
	
	Among baseline LMs, Alpaca, the only instruction
	tuned LM without reformulated NLP tasks, achieves
	the best overall performance. However, it still
	performs worse than our \ModelName models,
	potentially due to the noise in its training data,
	Self-Instruct. Furthermore, OPT-30B, despite having
	the highest number of parameters, fails to follow
	instructions as effectively as instruction-tuned \ModelName-T5-XL and \ModelName-OPT-2.7B, which are ten times smaller (see Figure \ref{fig:res_different_sizes}).
	This observation is consistent with prior findings that demonstrate the effectiveness of instruction tuning in smaller LMs \cite{ouyang2022training}.
		
	\subsection{Out-of-domain Generalization}

\begin{table}[t]
	\setlength\tabcolsep{4.6pt}
	\centering
	\resizebox{\columnwidth}{!}{
		\begin{tabular}{lrr@{\hspace{0.0cm}}rrr}
			\toprule
			{} &  \textbf{Avg.} &  \textbf{RGen} &&  \textbf{ELI5} &  \textbf{WP} \\
			\midrule
			T0++     &                    10.9 &                     18.7 &&                     3.8 &                   10.2 \\
			T\textit{k}-Instruct      &                     6.3 &                     12.9&\textsuperscript{\textdagger} &                     3.6 &                    2.4 \\
			Flan-T5 &                    10.6 &                     20.9&\textsuperscript{\textdagger}	 &                     3.5 &                    7.4 \\
			Alpaca-LLaMA-7B &                14.6 &                     19.5 &&                    12.5 &                   11.8 \\
			\midrule
			OPT-30B  &                    11.1 &                     18.6 &   &                 12.2 &                    2.6 \\
			
			\midrule
			\ModelName-T5-XL  &                    16.3 &                     20.2 &&                    18.3 &                   10.6 \\
			\ModelName-OPT-6.7B   &                    17.7 &                     16.9& &                    17.2 &                   \textbf{19.0} \\
			\ModelName-LLaMA-7B &                    \textbf{19.7} &                     \textbf{21.7} &&                    \textbf{18.6} &                   18.9 \\
			\bottomrule
		\end{tabular}
	}
	\caption{METEOR scores for out-of-domain
		datasets.  \ModelName models
		outperform prior instruction-tuned
		LMs
		on  recipe generation (RGen), long-form question answering
		(ELI5) and short story generation (WP).
		\textdagger{}: T\textit{k}-Instruct
		and FLAN-T5 are trained on RGen.}
	\label{tab:res_generalization}
\end{table}
	
To evaluate the generalization capability
	of \ModelName models, we evaluate on
	three diverse datasets: recipe generation
        (RGen) \cite{bien-etal-2020-recipenlg},
	long form question answering
	(ELI5) \cite{fan-etal-2019-eli5}, and short story
	generation
	(WP). We note  that
	the \ModelName models are not trained on any of these three datasets while the training set of T\textit{k}-Instruct and Flan-T5 includes RGen.

		Table \ref{tab:res_generalization}
shows that the \ModelName models consistently outperform
baseline LMs in all tasks, with \ModelName-LLaMA-7B
achieving a 35\% higher relative METEOR score than the best
baseline LM, Alpaca-LLaMA-7B. While they both have the
same underlying LM, our LM achieves higher
scores than Alpaca. This can be attributed to the diverse
and human-written nature of \DataName compared to
Alpaca's fully synthetic examples.

We present qualitative examples from the best
performing \ModelName model and the best baseline LM,
Alpaca,
in
Figure \ref{fig:qual_ex_generalizability} in the Appendix
(\S\ref{sec:qualitative_examples}). This
illustrates that \ModelName models have better factual
accuracy, instruction-following, long form text planning
than
prior instruction-tuned LMs.

\begin{table}[t]
	\setlength\tabcolsep{2.5pt}
	\centering
	\resizebox{\columnwidth}{!}{
		\begin{tabular}{llrr}
			\toprule
			\textbf{Model} & \textbf{Dataset} & \textbf{NLG} &  \textbf{NLU (MMLU)} \\
			\midrule
			
			\multirow{3}{*}{LLaMA-7B} & FLAN &                           9.1 &                     36.6 \\
			&      \DataName     		&                                  \textbf{19.7} &                     35.2 \\
			\cmidrule(lr){2-4}
			&   FLAN+\DataName	  &  											  16.5 &        \textbf{38.9} \\
			\midrule
			\midrule
			\multirow{3}{*}{LLaMA 2-7B} & FLAN &               13.3      &                     43.5 \\
			&      \DataName     		&                                        \textbf{19.3}           &                44.7  \\
			\cmidrule(lr){2-4}
			&   FLAN+\DataName	  &  							17.8  &  \textbf{46.2} \\
			\midrule
			\midrule
			\multirow{3}{*}{Mistral 7B} & FLAN &        18.9                    &             57.7 \\
			&      \DataName     		&                          \textbf{22.0}    &               \textbf{58.8}   \\
			\cmidrule(lr){2-4}
			&   FLAN+\DataName	  &  						19.4			 &  58.3   \\
			\bottomrule
		\end{tabular}
	}
	\caption{\DataName{} consistently outperforms FLAN on NLG tasks (RGen, ELI5, WP) across all models. For LLaMA-7B,
		FLAN+\DataName{} achieves the best MMLU score, while \DataName{} 
		outperforms FLAN in both NLG and NLU with LLaMA 2-7B and Mistral 7B: the improvements with \methodnames 
		are consistent and even more powerful with stronger models.}
	
	\label{tab:nlu_vs_nlg}
\end{table}

	\subsection{Language Understanding and Generation}

	We evaluate how much \DataName{} improves natural 
	language understanding (NLU) in addition to NLG. To
	this end,
	we compare \DataName{} with the strong baseline
 FLAN 
	\cite{longpre2023flancollection}, which
	performs well on NLU.
	We finetune base LMs using \DataName{}, 
	FLAN and their combination, 
	evaluating performance on OOD NLG 
	(i.e., recipe generation, ELI5 question answering, 
	and story generation) and 5-shot MMLU.
	We extend our evaluation to
	newer, more powerful models than LLaMA-7B: LLaMA 2-7B
	and Mistral 7B.

	Table \ref{tab:nlu_vs_nlg} shows that LLaMA-7B finetuned with
	\DataName{} outperforms the FLAN baseline in NLG (+10.6).
	Incorporating \DataName{} with FLAN achieves the
	best MMLU performance (+2.3), though the combination
	gets lower NLG scores than \DataName{} alone (-3.2).
	
	\noindent \textbf{Extending LLMs:} Our analysis to more recent
	 LLMs shows promising results. We observe that,
	 with stronger models, \DataName 
	 shows better results than FLAN (both in NLU and NLG).
	  With LLaMA 2-7B, \DataName{} again improves 
	 NLG (+6.0) while narrowing the gap in NLU, even surpassing 
	 FLAN's MMLU score (+1.2). The FLAN+\DataName{} combination further 
	 improves MMLU  (+2.7). With Mistral 7B, \DataName{} 
	 outperforms both FLAN and the combination,
	not only on NLG
	(+3.1, +2.6) but also on MMLU (+1.1, +0.5).
	These findings indicate that \DataName{} consistently improves NLG performance 
	across LLMs, with 
	its impact on NLU becoming more pronounced in stronger LLMs. This suggests that
	more natural datasets
	like \DataName{} may lead to better improvements in stronger LLMs than 
	NLP reformulation datasets
	like FLAN.

	\subsection{Ablation of \DataName{} Subsets}
We now conduct an ablation study to investigate the
differential impact of the three distinct subsets of 
	\DataName{} by finetuning LLaMA-7B: NLP Tasks
	(NLP), Structured Corpora (SC), and \MethodNames
	(RI).

\begin{table}[t]
	\setlength\tabcolsep{4.6pt}
	\centering
	\resizebox{\columnwidth}{!}{
		\begin{tabular}{llrr}
			\toprule
			\textbf{Model} & \textbf{Subset} & \textbf{NLG} &  \textbf{NLU (MMLU)} \\
			\midrule
			
			\multirow{4}{*}{LLaMA-7B} & NLP &               13.6  &                     35.4 \\
			&      NLP+SC     		&                                  18.6  &                     35.0 \\
			&      NLP+RI     		&                                  \textbf{19.9} &                     \textbf{35.5} \\
			\cmidrule(lr){2-4}
			&   NLP+RI+SC	  &  											  19.7 &        35.2 \\
			
			\bottomrule
		\end{tabular}
	}
	\caption{Subset ablation experiments to study the effect of different subsets in \DataName. Adding the \MethodNames{} subset further improves the performance both on NLG and NLU showing its capabilities while NLP and SC subsets show limited improvements. }
	\label{tab:ablation_study}
\end{table}
	
We summarize the results in
Table \ref{tab:ablation_study}. Since the NLP tasks subset
already includes examples with long form output, the model
performs comparatively well on NLG, achieving better
performance than the performance of FLAN
(see Table \ref{tab:nlu_vs_nlg}), another NLP tasks
dataset for instruction-tuning. We observe a trade-off between
NLG (+5.0) and NLU (-0.4) performance when
we include the SC subset in addition to NLP tasks. However,
including the \MethodNames{} subset significantly improves
NLG (+6.3) and also slightly improves
NLU (+0.1). Despite the instructions
in the SC subset being created directly from human
annotation (StackExchange) or via minimal changes (WikiHow), 
the improvement by \MethodNames{} is
substantially better and more consistent. We achieve the
best results when the \MethodNames{} (RI) subset is
included.

\subsection{Data Evaluation}
Our instruction generation (for \methodnames) relies
on LLM generation. To ensure its quality, we conduct a human evaluation. 
We randomly select 100 corpus examples 
from \DataName
(without adding length
information) and provide them to Mechanical Turk
annotators. Annotators are asked to determine
whether the generated instruction is relevant to the
given document or not. 97 out of 100 instructions
are found to be relevant. This high degree of relevance 
suggests that combining instruction generation with 
corpus examples can potentially improve the effectiveness 
of LLMs for instruction-following compared to the 
Self-Instruct dataset \cite{wang2022selfinstruct}, 
which relies on LLMs to generate tasks, instructions, 
and outputs, and achieves a validity of 54\%.

We conducted an additional manual analysis of a corpus
subset of \DataName  to identify areas
for improvement in corpus mining and instruction tuning,
providing valuable insights for future work. We present
qualitative examples in Table \ref{tab:selected_failures} in
the Appendix. Our analysis shows that the failure cases
mostly stem from the detailed nature of  corpus texts
such as news. While \methodnames is capable of generating
relevant instructions, the corpus examples can sometimes
include extensive content. This tendency can lead \ModelName
models to generate long responses even for simple questions
that require concise answers.  This issue occurred
infrequently, as supported by the overall performance
of \ModelName.  However, this is still an important insight
for future work in this area.

	\section{Conclusion}
	In this paper, we introduce \methodnames, a novel method 
	to transform human-written texts into instruction-output pairs
	via LLMs. We present \DataName, an
	instruction-following long text generation dataset
	that combines \methodnames examples, 
	structured corpora (SC), and NLP tasks. Our evaluation shows that
	the generated reverse instructions are highly relevant to
	the corpus examples and contain a diverse set of
	tasks. Furthermore, we demonstrate that
	our \ModelName models -- instruction-tuned on
	\DataName  -- outperform prior
	instruction-tuned baselines such as FLAN-T5 and
	Alpaca by a big margin on a wide variety of
        long text generation tasks.
	
	\section*{Ethics Statement}
	\ModelName models have capabilities of generating high-quality long texts, such as news articles, essays, and tutorials by following given instructions. Therefore, this may raise concerns regarding the potential for malicious use. However, we recognize the importance of publicly sharing our models and datasets to facilitate further research. By making \ModelName models easily accessible, we hope to support researchers to explore potential implications of instruction-following large language models, including techniques for improving their truthfulness or watermarking model outputs. Therefore, we believe that the possible benefits of \ModelName models for the research community outweigh their possible risks.
	
	\section*{Limitations}
	The proposed datasets and models mainly focus on long text generation and may have limitations regarding structured prediction tasks in NLP with some LLMs. We observe that \DataName{} is a good complementary dataset and when combined with existing datasets, it further improves NLU performance. However, for more recent models, it outperforms previous datasets not only in NLG but also on structured prediction tasks. Additionally, we observe that \ModelName{} models may present hallucination problems similar to those found in LLMs.

	\section*{Acknowledgements}
	This work was funded by Deutsche Forschungsgemeinschaft
	(project SCHU 2246/14-1).

	\bibliography{anthology,custom}
	\bibliographystyle{acl_natbib}
	\newpage
	\appendix
	\section{Training Details}
\label{sec:training_details}
We include an end-of-instruction token, [EOI], between the
instruction and the output of the autoregressive
LMs, OPT
and LLaMA variants, to differentiate between  instruction
and output. We employ two to four A6000 48GB GPUs for
finetuning  with a batch size of 32 to 64,
and train with DeepSpeed%
, including bfloat16
training. The learning rate is determined by comparing the
minimum validation loss during three epochs:  5e-5
for \ModelName-T5-3B, 5e-6 for \ModelName-OPT-6.7B
and \ModelName-LLaMA-7B. Apart from the three large
LMs, we finetune four smaller OPT LMs (125M, 350M,
1.3B, 2.7B) with \DataName and  release them
to increase accessibility of instruction-tuned
LMs. See \S\ref{sec:smaller_longform} for more details.

\section{Evaluation with Different Metrics}
\label{sec:appendix_results}
We extend our evaluation setup with BLEU, ROUGE, and self-BLEU in addition to the METEOR metric to show more robust results. The results indicate that we have similar patterns in all parts of evaluation. We report average in-domain performance in Table \ref{tab:extended_domain_results}, out-of-domain performance in Table \ref{tab:extended_ood_results}, and multilingual performance in Table \ref{tab:extended_multilingual_results}.

\begin{table*}[tp]
	\centering
	\begin{tabular}{lrrrr}
		\toprule
		&    \textbf{METEOR} &    \textbf{BLEU} &    \textbf{ROUGE-L} &   \textbf{self-BLEU} \\
		\midrule
		T0++ 						         &    5.9 &  0.0 &  8.6 &  0.02 \\
		T\textit{k}-Instruct 	      &    6.0 &  0.0 &  9.2 &  0.02 \\
		Flan-T5 				  	 	    &  12.5 &  0.1 &  16.0 &  0.04 \\
		Alpaca-LLaMA-7B 		  &  15.2 &  1.8 &  20.4 &  0.10 \\
		\midrule
		OPT-30B 						 &  12.3 &  2.1 &  13.4 &  0.18 \\		
		\midrule
		\ModelName-T5-XL  		&  24.3 &  \textbf{9.4} &  26.0 &  \textbf{0.19} \\
		\ModelName-OPT-6.7B  &  24.2 &  8.8 &  27.1 &  \textbf{0.19} \\
		\ModelName-LLaMA-7B &  \textbf{24.8} &  8.8 &  \textbf{27.6} &  0.16 \\
		\bottomrule
	\end{tabular}
	\caption{Average in-domain performance of baseline and \ModelName models with METEOR, BLEU, ROUGE, and self-BLEU. METEOR scores of each subset are reported in the Table \ref{tab:res_writinginstructions}.}
	\label{tab:extended_domain_results}
	
\end{table*}

\begin{table*}[tp]
	\centering
	\begin{tabular}{lrrrr}
		\toprule
		&    \textbf{METEOR} &    \textbf{BLEU} &    \textbf{ROUGE-L} &   \textbf{self-BLEU} \\
		\midrule
		T0++ 						         &   10.9 &  0.3 & 14.4 &  0.07 \\
		T\textit{k}-Instruct 	      &    6.3 &  0.0 &  9.4  &  0.04 \\
		Flan-T5 				  	 	    &  10.6 &  0.2 &  14.5 &  0.07 \\
		Alpaca-LLaMA-7B 		  &  14.6 &  1.5 &  \textbf{22.1} &  0.13 \\
		\midrule
		OPT-30B 						 &  11.1  &   0.8 &  11.6 &  0.12 \\		
		\midrule
		\ModelName-T5-XL  		&  16.3 &  1.7 &  16.2 &  0.22 \\
		\ModelName-OPT-6.7B  &  17.7 &  1.4 &  17.4 &  \textbf{0.27} \\
		\ModelName-LLaMA-7B &  \textbf{19.7} &  \textbf{2.1} &  21.6 &  0.20 \\
		\bottomrule
	\end{tabular}
	\caption{Average out-of-domain performance of baseline and \ModelName models with METEOR, BLEU, ROUGE, and self-BLEU. METEOR scores of each subset are reported in the Table \ref{tab:res_generalization}.}
	\label{tab:extended_ood_results}
	
\end{table*}

\begin{table*}[tp]
	\centering
	\begin{tabular}{lrrrr}
		\toprule
		&    \textbf{METEOR} &    \textbf{BLEU} &    \textbf{ROUGE-L} &   \textbf{self-BLEU} \\
		\midrule
		T0++ 						         &    1.8 &  0.0 &  1.8 &  0.12 \\
		T\textit{k}-Instruct 	      &    0.8 &  0.0 &  1.9 &  0.01 \\
		Flan-T5 				  	 	    &   2.4 &  0.0 &  4.3 &  0.08 \\
		Alpaca-LLaMA-7B 		  &   4.9 &  0.1 &  7.7 &  0.14 \\
		\midrule
		OPT-30B 						 &   4.5 &  0.0 &  5.8 &  0.07 \\		
		\midrule
		\ModelName-T5-XL  		&   6.9 &  0.4 &  5.8 &  \textbf{0.39} \\
		\ModelName-OPT-6.7B  &   9.5 &  0.3 &  10.5 &  0.17 \\
		\ModelName-LLaMA-7B &   \textbf{9.7} &  \textbf{0.6} &  \textbf{10.7} &  0.18 \\
		\bottomrule
	\end{tabular}
	\caption{Average multilingual performance of baseline and \ModelName models with METEOR, BLEU, ROUGE, and self-BLEU.  METEOR scores of each language are reported in the Table \ref{tab:res_multilingual}.}
	\label{tab:extended_multilingual_results}
	
\end{table*}

\begin{table}[t]
	\footnotesize
	\setlength\tabcolsep{4.6pt}
	\centering
	\begin{tabular}{lrrrrr}
		\toprule
		{} &  \textbf{Avg.} &  \textbf{DE} &  \textbf{ES} &  \textbf{FR} &  \textbf{RU} \\
		\midrule
		T0++    &                     1.8 &                    1.6 &                    2.4 &                    2.5 &                    0.7 \\
		T\textit{k}-Instruct     &                     0.8 &                    0.8 &                    1.1 &                    1.0 &                    0.5 \\
		Flan-T5 &                     2.4 &                    2.5 &                    2.8 &                    3.4 &                    0.7 \\
		Alpaca-LLaMA-7B &                     4.9 &                    7.9 &                    7.0 &                    4.1 &                    0.7 \\
		\midrule
		OPT-30B  &                     4.5 &                   \textbf{12.7} &                    3.5 &                    1.3 &                    0.5 \\
		\midrule
		\ModelName-T5-XL   &                     6.9 &                    5.2 &                    9.0 &                    9.1 &                    4.2 \\
		\ModelName-OPT-6.7B   &                     9.5 &                    7.3 &                   \textbf{12.2} &                   \textbf{15.3} &                    3.1 \\
		\ModelName-LLaMA-7B &                     \textbf{9.7} &                    8.9 &                   10.5 &                   14.8 &                   \textbf{4.7} \\
		\bottomrule
	\end{tabular}
	\caption{Evaluation of multilingual news generation via METEOR.}
	\label{tab:res_multilingual}
\end{table}

\section{Multilingual News Generation}
\label{sec:appendix_multilingual}
We evaluate the multilingual news generation
capabilities of \ModelName models against the
baselines. Although all LMs in our
experiments are pretrained and instruction-tuned
mainly on English, their pretraining data  includes
other languages.\footnote{20
	languages from Wikipedia are included in LLaMA, which accounts for 4.5\%
	of the entire pretraining data.}
Table \ref{tab:res_multilingual} shows
that \ModelName models outperform baselines
on the task of generating a news article given a title
in Spanish, French, and Russian.

We observe that \ModelName-LLaMA-7B shows the best performance
in the multilingual setup as well. We also observe that all
OPT variants (\ModelName-OPTs and OPT-30B) have a
higher rate of generating text in the given language
(see Table \ref{tab:lang_prediction} in the
Appendix), which results in better performance for OPT.
In cases where the LM generates
English text while the instruction is in a different
language, our manual inspection shows
that \ModelName models can still follow the non-English
instruction. Additionally, when we generate with
different seeds, we observe that \ModelName models are
able to generate in the language of the
instruction. We present qualitative multilingual
examples in Figure \ref{fig:qual_ex_multilingual} in
the Appendix. Finally, we believe that our
methodology (the use of
corpus examples for instruction tuning data) is
extensible  to other languages, with less
reliance on
multilingually highly competent
LLMs.

We observe that while we ask instructions in different languages, the models might have tendencies to respond in English. We present the distribution of outputs in the language that is given in the instruction in Table \ref{tab:lang_prediction}, detected by the language-detection library  \cite{nakatani2010langdetect}. We see that raw language model, OPT-30B, and \ModelName-OPT-6.7B can respond with higher ratio in the given language. While Flan-T5 and T\textit{k}-Instruct have higher ratio of generating text in the language of instruction, it is important to note that the quality of the generated text is much lower as it can be seen in Table \ref{tab:res_multilingual}.
\begin{table}[H]
	\footnotesize
	\setlength\tabcolsep{4.6pt}
	\centering
	\begin{tabular}{lrrrr}
		\toprule
		&    \textbf{DE} &    \textbf{ES} &    \textbf{FR} &   \textbf{RU} \\
		\midrule
		T0++ &  0.09 &  0.12 &  0.07 &  0.56 \\
		T\textit{k}-Instruct &  0.96 &  0.95 &  0.97 &  0.55 \\
		Flan-T5 &  0.98 &  0.95 &  0.99 &  0.45 \\
		Alpaca-LLaMA-7B &  0.89 &  0.90 &  0.26 &  0.23 \\
		\midrule
		OPT-30B &  0.96 &  0.99 &  1.00 &  1.00 \\		
		\midrule
		\ModelName-T5-XL &  0.35 &  0.74 &  0.57 &  0.89 \\
		\ModelName-OPT-6.7B &  0.50 &  0.88 &  0.94 &  1.00 \\
		\ModelName-LLaMA-7B &  0.77 &  0.91 &  0.93 &  0.44 \\
		\bottomrule
	\end{tabular}
	\caption{The ratio of generating text in the language of the instruction.}
	\label{tab:lang_prediction}
	
\end{table}

\section{Smaller \ModelName Models}
\label{sec:smaller_longform}
\begin{figure}[H]
	\centering
	\includegraphics[width=0.99\linewidth]{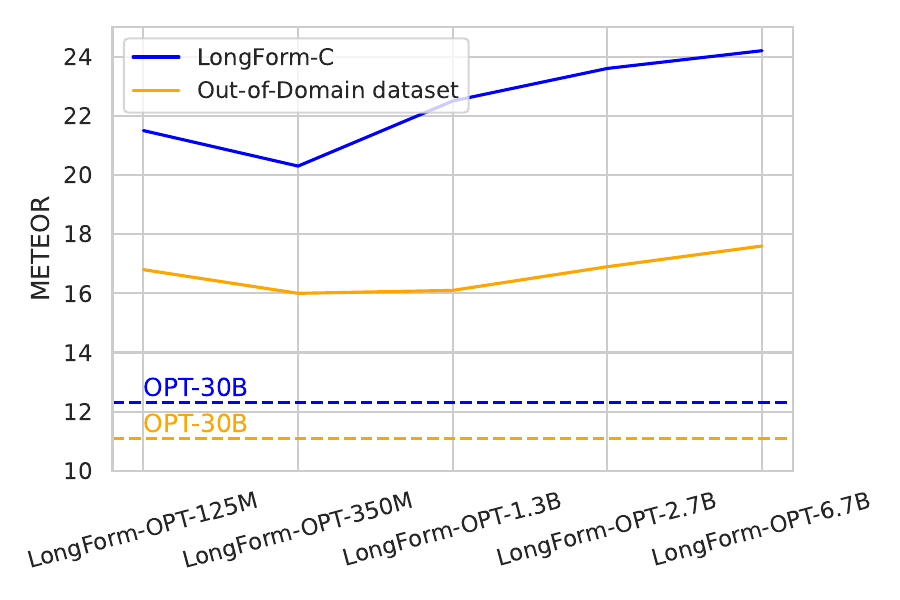}
	\caption{Comparing METEOR scores for varying sizes of \ModelName-OPT models on in-domain (\DataName) and out-of-domain (RGen, ELI5, WP) tasks. All instruction-tuned \ModelName-OPT models outperform OPT-30B (dashed) without instruction tuning.}
	\label{fig:res_different_sizes}
\end{figure}

In order to increase accessibility, we finetune varying size of PLMs, OPT-125M, OPT-350M, OPT-1.3B, OPT-2.7B, and OPT-6.7B models with the \DataName dataset. We see that all \ModelName-OPT models have better METEOR scores than OPT-30B without instruction tuning. We present average METEOR scores of \ModelName-OPT models and OPT-30B in Figure \ref{fig:res_different_sizes}.

\section{Qualitative Examples}
\label{sec:qualitative_examples}
We present qualitative examples from the best performing \ModelName model and the best baseline model, Alpaca, for each task in Figure \ref{fig:qual_ex_generalizability}.

\begin{figure*}[t]
	\centering
	\scriptsize
	\begin{tabular}{p{0.35\linewidth} p{0.60\linewidth}}
		\toprule
		\multicolumn{2}{l}{\textbf{Recipe Generation Instruction:} }  \\
		\multicolumn{2}{p{0.95\linewidth}}{
			Ingredients: beef, mushroom soup, mushrooms.
			
			Write directions of a cooking recipe with these ingredients.
		}\\
		\midrule
		\textbf{Alpaca-LLaMA-7B completion:} & \textbf{\ModelName-LLaMA-7B completion:} \\
		In a pot, add a can of \greenhl{mushroom soup} and 2 cups of \redhl{beef broth}. Heat to a simmer and add 3 cups of sliced \greenhl{mushrooms}. Let simmer for 15 minutes. Serve over cooked noodles or rice. Enjoy!
		&
		Put the \greenhl{meat} in a large saucepan or skillet and cover it with water. Add 1 tsp (5 ml) of salt and bring it to a boil.
		Remove the meat from the pan and discard the water.
		Add 1/2 can (125 ml) of \greenhl{mushroom soup} and 2 cups (473 ml) of \greenhl{mushrooms} and simmer it for 15 minutes.
		Sprinkle with salt and pepper and garnish with sour cream and parsley before serving.
		\\
		
		\midrule[\heavyrulewidth]
		
		\multicolumn{2}{l}{\textbf{ELI5 Instruction:} }  \\
		\multicolumn{2}{p{0.95\linewidth}}{
			how do digital cameras capture and transfer what they see into a .jpg?
		} \\
		\midrule
		\textbf{Alpaca-LLaMA-7B completion:} & \textbf{\ModelName-LLaMA-7B completion:} \\
		
		Digital cameras capture and transfer the light reflected from an object into a.jpg file using a process called \redhl{image compression}. The image compression algorithm takes a series of pixels from the image and arranges them into blocks of similar colors. It then reduces the amount of data needed to store the image by removing redundant information from the blocks. The final result is an.jpg file that can be stored, edited, and shared.
		&
		It doesn't matter how the camera captures the picture; what it captures is a signal, which is an analogue value. That signal is either converted to digital form by a linear \greenhl{analog-to-digital converter (ADC)}, usually at the image sensor itself, or it's first converted to a voltage which is then read in by a linear ADC, either at the image sensor or in the digital part of the image processing pipeline. The digital value of the ADC is converted by software into a.jpg.
		
		There's a lot of clever maths going on behind the scenes; but the crucial part is that it is a digital value which is converted to a.jpg.\\
		\midrule[\heavyrulewidth]
		
		\multicolumn{2}{l}{\textbf{WritingPrompts Instruction:} }  \\
		\multicolumn{2}{p{0.95\linewidth}}{
			Prompt:  The heroes have finally done it! They've overthrown the evil overlord, destabilized his government and saved the world! While celebrating they very suddenly realize that they have no idea how to run a government.
			Write a story about given prompt.
		}\\
		\midrule
		\textbf{Alpaca-LLaMA-7B completion:} & \textbf{\ModelName-LLaMA-7B completion:} \\
		It was a momentous occasion, a day the heroes would never forget. They had banded together to save the world from the evil overlord, and their triumph was complete. After the celebrations had died down, however, the heroes realized they had no idea how to run a government. The world needed a new leader, but who should it be? What decisions should they make? The heroes knew they were up to the task, but they needed to think carefully if they wanted to do it right. 
		
		\begin{minipage}{0.5\textwidth}{\tiny{\textit{(85 words)}}}\end{minipage}
		&
		It was over. The final battle, the last stand, and the victory celebration had concluded and the final tally was in: the villains were dead, their minions were scattered or captured, the last of the evil artifacts were destroyed, and the world had been saved.
		
		The cheering heroes took a break from their celebrations to hear the good news from the new king of the newly formed world government. He had good news: the world was once again at peace, all villain threats were eliminated, and a new era of prosperity was just beginning. All that remained were some minor issues of governance and organization. He was currently working on those and would be happy to hand over full control to the new rulers as soon as he finished.
		
		The heroes looked at each other with doubt and dismay. They were heroes, not politicians. They couldn’t run a world. They were terrible at that sort of thing.
		
		Then the realization struck. “You want us to be your government?” The new king was taken aback. “Yes. That’s why you’re here. To take over for me. All I need is your names and titles and I can hand over all power.” The heroes had a serious crisis. 
		
		\begin{minipage}{0.5\textwidth}{\tiny{\textit{(205 words)}}}\end{minipage}
		\\
		
		\bottomrule
	\end{tabular}
	\caption{Generalization capabilities of \ModelName{ }for three out-of-domain tasks:
          long form Q\&A, story and recipe generation. \ModelName{ }generates well-structured, reliable, and creative long texts compared to the best baseline model. In particular, Alpaca suffers from factual inaccuracies in ELI5, limited creativity in WritingPrompts with shorter response, and a tendency to ignore ingredients in RGen. Red highlights indicate where the model may be hallucinating or deviating from instructions while green highlights accurate generation.}	
	\label{fig:qual_ex_generalizability}
\end{figure*}

\section{Details of \DataName}
\subsection{LLM Query}
\label{sec:llm_query_appendix}

We use OpenAI's text-davinci-003 Completion API for our LLM query part. We just use the default parameters in which temperature and top\_p parameters are 1. 

The template for the instruction style:
\begin{mdframed}[hidealllines=true,backgroundcolor=gray!5]
	{\small
	Instruction: X  
	
	\noindent Output: \textit{<corpus\_example>} 
	
	\noindent What kind of instruction could this be the answer to?  
	
	\noindent X:}
\end{mdframed}

The template for the informal chatbot style:
\begin{mdframed}[hidealllines=true,backgroundcolor=gray!5]
	{\small
	You are a chatbot. A user sent you an informal message and your reply is as follows. 
	
	\noindent Message: X
	
	\noindent Reply: \textit{<corpus\_example>}
	
	\noindent What is the informal message X?
	
	\noindent X:}
\end{mdframed}

The template for the search engine/query style:
\begin{mdframed}[hidealllines=true,backgroundcolor=gray!5]
	{\small
	You are a search engine. A person queried something in detail and the most relevant document about the query is as follows.
	
	\noindent Query: X  
	
	\noindent Document: \textit{<corpus\_example>}
	
	\noindent What is the detailed query X?
	
	\noindent X:}
\end{mdframed}

\subsection{WikiHow Templates}
\label{sec:wikihow_templates_appendix}
Below, we list 18 WikiHow templates utilized in creating the \DataName. These templates feature questions in the form of verb phrases, for example, ``make conversation'' in addition to number of steps.

{\small
	\begin{enumerate}
		\itemsep0em 
		
		\item Give me [STEP] steps to [QUESTION].
		\item How to [QUESTION]?
		\item Do you know how can I [QUESTION]?
		\item List [STEP] instructions to [QUESTION].
		\item What are some tips to [QUESTION]?
		\item What are some steps to [QUESTION]?
		\item Can you provide [STEP] clear and concise instructions on how to [QUESTION]?
		\item I'm interested in learning how to [QUESTION]. Could you break it down into [STEP] easy-to-follow steps?
		\item For someone who is new to [QUESTION], what would be [STEP] key steps to get started?
		\item What is the most efficient way to [QUESTION]? Could you provide a list of [STEP] steps?
		\item Do you have any advice on how to [QUESTION] successfully? Maybe a step-by-step guide with [STEP] steps?
		\item I'm trying to accomplish [QUESTION]. Could you walk me through the process with [STEP] detailed instructions?
		\item What are the essential [STEP] steps to [QUESTION]?
		\item I need to [QUESTION], but I'm not sure where to start. Can you give me [STEP] actionable steps?
		\item As a beginner in [QUESTION], what are the [STEP] basic steps I should take?
		\item I'm looking for a comprehensive guide on how to [QUESTION]. Can you provide [STEP] detailed steps?
		\item Could you outline [STEP] practical steps to achieve [QUESTION]?
		\item What are the [STEP] fundamental steps to consider when attempting to [QUESTION]?
	\end{enumerate}
}

\subsection{Enron Templates}
\label{sec:enron_templates_appendix}
We list the instructions for email writing task for the given subject.
{\small
	\begin{enumerate}
		\itemsep0em 
		\item Write an email with the subject "[SUBJ]"
		\item Can you craft an email with the subject [SUBJ]?
		\item Would you be able to compose an email and use [SUBJ] as the subject?
		\item Create an email about [SUBJ].
		\item Draft an email and include the subject "[SUBJ]".
		\item Generate an email about [SUBJ].
		\item Hey, can you shoot me an email about [SUBJ]?
		\item Do you mind crafting an email for me with [SUBJ] as the subject?
		\item Can you whip up an email with the subject of "[SUBJ]"?
		\item Hey, can you write an email and use "[SUBJ]" as the subject?
		\item Can you send me an email about [SUBJ]?
	\end{enumerate}
}
\subsection{BEA-GEC Templates}
\label{sec:bea_gec_templates_appendix}
We list the instructions for grammar error correction task. We prepend each input to randomly selected instruction.
{\small
	\begin{enumerate}
		\itemsep0em 
		\item Edit and revise this document to improve its grammar, vocabulary, spelling, and style.
		\item Revise this document to correct all the errors related to grammar, spelling, and style.
		\item Refine this document by eliminating all grammatical, lexical, and orthographic errors and improving its writing style.
		\item Polish this document by rectifying all errors related to grammar, vocabulary, and writing style.
		\item Enhance this document by correcting all the grammar errors and style issues, and improving its overall quality.
		\item Rewrite this document by fixing all grammatical, lexical and orthographic errors
		\item Fix all grammar errors and style issues and rewrite this document
		\item Take a stab at fixing all the mistakes in this document and make it sound better.
		\item Give this document a once-over and clean up any grammar or spelling errors.
		\item Tweak this document to make it read smoother and fix any mistakes you see.
		\item Make this document sound better by fixing all the grammar, spelling, and style issues.
		\item Proofread this document and fix any errors that make it sound weird or confusing.
	\end{enumerate}
}

\section{Diversity Analysis: LongForm-C, Dolly, FLAN}
\label{sec:diversity_comparison}

\begin{figure*}[h]
	\centering
	\begin{subfigure}{0.3\textwidth}
		\includegraphics[width=\textwidth]{Figures/pie_chart_longform.pdf}
		\caption{\DataName}
		\label{fig:instruction_analysis_ours}
	\end{subfigure}
	\begin{subfigure}{0.3\textwidth}
		\includegraphics[width=\textwidth]{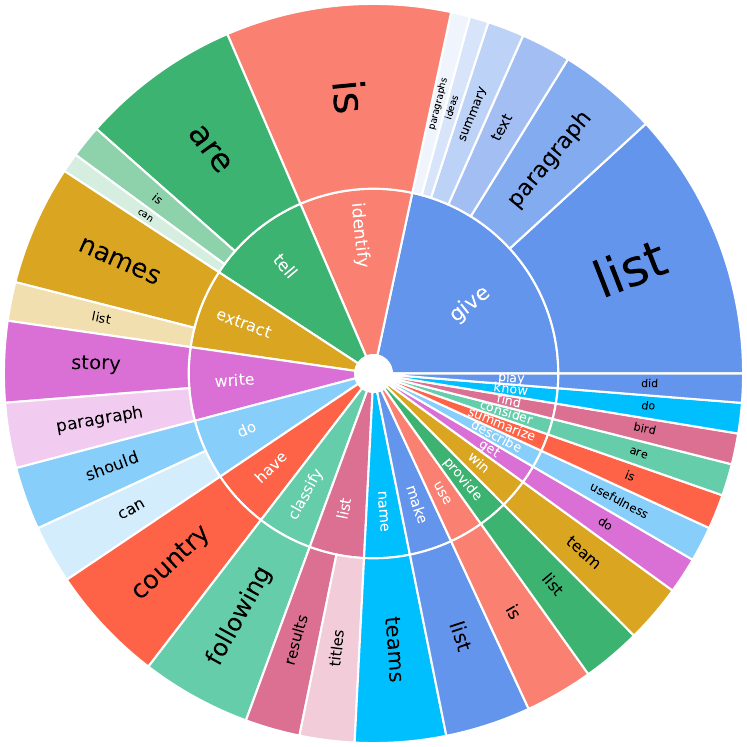}
		\caption{Dolly}
		\label{fig:instruction_analysis_dolly}
	\end{subfigure}
	\begin{subfigure}{0.3\textwidth}
		\includegraphics[width=\textwidth]{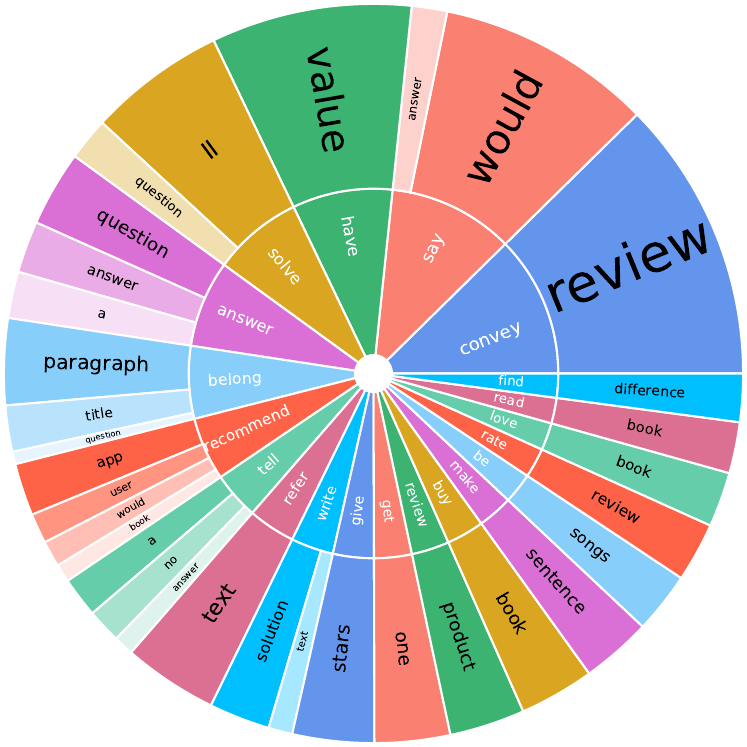}
		\caption{FLAN}
		\label{fig:instruction_analysis_flan}
	\end{subfigure}
	\caption{Comparison of instruction diversity across \DataName, Dolly, and FLAN datasets. Each subfigure shows the distribution of noun+verb and auxiliary+verb pairs extracted from instructions.}
	\label{fig:combined_instruction_analysis}
\end{figure*}

To further analyze the diversity of instructions in \DataName, we compare its distribution of noun+verb and auxiliary+verb pairs (Figure \ref{fig:instruction_analysis_ours}) with those of the Dolly human annotation dataset (Figure \ref{fig:instruction_analysis_dolly}) and the FLAN dataset (Figure \ref{fig:instruction_analysis_flan}), which focuses specifically on NLP tasks. Figure \ref{fig:combined_instruction_analysis} presents these three distributions side-by-side for easy comparison.

We observe that \DataName exhibits a variety of instruction types, covering areas such as description, explanation, analysis, and creative tasks. In contrast, the Dolly dataset appears more focused on listing, naming, extraction, and identifying tasks, while still maintaining diversity. The FLAN dataset, being NLP-specific, shows a concentration around text classification, question answering, and sentiment analysis tasks.

This comparison highlights how \methodnames in \DataName generate diverse tasks that complement existing datasets like Dolly and FLAN, suggesting potential benefits in combining multiple data collection strategies for more comprehensive coverage.

\section{Manual Analysis of the \DataName Dataset}
We conduct a manual analysis of the \methodnames subset of the \DataName subset. We present qualitative examples in Table \ref{tab:selected_failures}.
\begin{table*}[t]
	\centering
	\scriptsize
	\begin{tabular}{p{0.15\linewidth} p{0.80\linewidth}}
		\toprule
		\textbf{Generated Instruction:} & \textbf{Corpus Example:} \\
		Describe a new research study on brain biometrics.
		&
		\redhl{\parbox{\linewidth}{Credit: Chen Song, University at Buffalo.
		BUFFALO, N.Y. — To overcome password fatigue, many smartphones include facial recognition, fingerprint scans and other biometric systems.}}
		``To the best of our knowledge, this is the first in-depth research study on a truly cancelable brain biometric system. We refer to this as ‘hard cancellation,’ meaning the original brain password can be reset without divulging the user’s identity,'' says collaborator Zhanpeng Jin, PhD, associate professor of computer science and engineering at UB.
		
		Their work is described in a study that will be presented June 11 at MobiSys 2018, a flagship mobile computing conference hosted in Germany by the Association for Computing Machinery. For a copy of the study, \redhl{please email Cory Nealon at cmnealon@buffalo.edu.}
		
		\redhl{\parbox{\linewidth}{The research was supported by the National Science Foundation (NSF) and NSF’s Center for Identification Technology Research.}}\\
		\midrule
		\multicolumn{2}{l}{\parbox{0.95\linewidth}{\textbf{Notes:} The output contains metadata information such as credit and location. It also includes title, contact information, and acknowledgments part which usually would not appear in instruction tuning dataset. Overall, this corpus example is not suitable for instruction tuning.}
		}  \\
		
		\midrule[\heavyrulewidth]
		\textbf{Generated Instruction:} & \textbf{Corpus Example:} \\
		Can you tell me if Scotland exists?
		&
		"Scotland: it certainly does exist. I can say that safely. \redhl{That mangling of 'Ypres' ↑ does nothing for my credibility amongst}
		
		\redhl{French-speakers though.}
		
		\redhl{Tagged with BlackMill Games, M2H, Verdun.''}\\
		\midrule
		\multicolumn{2}{l}{\parbox{0.95\linewidth}{\textbf{Notes:} This example consists of three distinct sentences. The first sentence is about Scotland, where the instruction is generated for. The second sentence mentions 'Ypres' because it appears in the title of the original context, which is truncated in the C4 dataset. The final sentence contains the tag information of this article, which has not been removed.}
		}  \\
		
		\midrule[\heavyrulewidth]
		\textbf{Generated Instruction:} & \textbf{Corpus Example:} \\
		What does Disney's streaming video service mean for Netflix?
		&
		Could the Mouses’s gauntlet leave Netflix bruised?
		
		That’s what investors are wondering after Disney announced it was not only going to launch its own streaming video service, but also no longer license its popular content to Netflix.
		
		The Disney-Netflix deal is slated to end in 2019. At that time, Disney plans to have its own subscription-based service that could cause Netflix subscribers to ditch their \$9.99 monthly fee.
		
		And according to a new survey from Fluent, as many as 1 in 5 of U.S.-based subscribers could leave the streaming behemoth. The survey shows that many Netflix users would at least consider cancelling their subscription if Disney pulls its content.
		According to the survey of 2,893 US adults aged 18 and up, 19 percent of total Netflix subscribers would stop their Netflix subscription if they lose access to Disney’s movies and content; 15 percent of non-parents said they would cancel their subscription; 23 percent of parents of kids ages 0-18 would cancel.
		
		What does all of this mean for investors? Potentially a devastating loss in revenue for Netflix.
		
		On average, Netflix subscribers pay \$10 per month. If 19 percent of the estimated 50.85 million U.S. subscribers cancel their service, Netflix could lose about \$96.6 million in monthly revenue, or about \$1.2 billion for the year.
		
		If Disney proves to be successful, consumers could expect other studios to follow suit, which could either result in Netflix losing subscribers, or those studios struggling to attract new subscribers. According to the survey, 62 percent of Americans aren’t likely to subscribe to an additional service, with cost likely being the driving factor. Still, 4 in 10 Americans are willing to pay for additional streaming services.
		
		Of course, Netflix isn’t just going to lie down and let this happen. The streaming giant knows content is king, and has been spending billions of dollars developing its own original series that attract subscribers. The company has nearly \$16 billion earmarked to be spent on content deals in the coming years, and it’s still adding subscribers at an impressive rate, recently exceeding 100 million global subscribers.
		
		One question that remains unclear is what happens to the Marvel content. Netflix is in “active discussions” with Disney about keeping Marvel and Star Wars films after 2019 when Disney and Pixar movies won’t be in the Netflix catalog, according to a Reuters report.
		
		\redhl{\parbox{\linewidth}{Andrew Dodson is a journalist who has worked in Michigan and Colorado. Email him at [email protected] Disclosure: Streaming Observer is supported by readers. Articles may contain referral links. For more information, see the disclosure at the bottom of the page.}}
		\\
		\midrule
		\multicolumn{2}{l}{\parbox{0.95\linewidth}{\textbf{Notes:} This is a news article about Disney's streaming service. Although it includes irrelevant details such as the reporter's name, the main issue is that the corpus example contains overly detailed answer. Usually, for instruction-tuned models, concise responses are preferred for such instructions.}
		}  \\
		
		\bottomrule
	\end{tabular}
	\caption{We present representative instances of failure cases in the corpus examples of the \DataName dataset. Although these examples are rare, we present them to guide future work in implementing additional filtering. The majority of these failure cases involve news articles or blog posts that contain irrelevant metadata information or are written in a detailed manner that may not be suitable for instruction tuning dataset.}	
	\label{tab:selected_failures}
\end{table*}

\newpage
\section{Qualitative Examples}
We present several qualitative examples from \ModelName that demonstrate its proficiency in trip planning, long-form writing, scriptwriting, and open-book QA.

\begin{figure*}[t]
	\centering
	\small
	\begin{tabular}{p{0.98\linewidth}}
		\toprule
		\multicolumn{1}{l}{\textbf{English Instruction:} }  \\
		\multicolumn{1}{p{0.98\linewidth}}{
			Write a blog post on the benefits of meditation. 
		}  \\
		\midrule
		\textbf{\ModelName-LLaMA-7B:} \\
		It doesn't matter where you are in the world, how old you are, or how much you know about meditation, it's for everyone. The benefits of meditation are endless.
		
		Meditation can be as simple as sitting quietly for five minutes with your eyes closed and focusing on your breath. The idea is to train your brain to let go of distracting thoughts and tension so you can think more clearly, feel more relaxed, and be more positive.
		
		Studies have shown that just a few minutes of meditation a day can lower your heart rate, decrease stress hormones, and even change your brain. [...] %
		\\
		\midrule[\heavyrulewidth]
		\multicolumn{1}{l}{\textbf{German Instruction:} }  \\
		\multicolumn{1}{p{0.98\linewidth}}{
			Schreiben Sie einen Blogbeitrag über die Vorteile der Meditation.
		}  \\
		\midrule
		\textbf{\ModelName-LLaMA-7B:} \\
		Meditation ist die einfachste und gleichzeitig geistigste Methode, um sich selbst zu entspannen und den unbewussten Gedankenkreis zu überprüfen. Nicht nur der Gedanke, sondern auch der Gefühlswelten, der Sensationen und Gefühle, die er hinterlässt, können durch ein regelmäßiges Meditationspraktizieren gelöst oder unter Kontrolle gebracht werden. Es gibt verschiedene Arten von Meditation, darunter die Konzentration auf einen Punkt, die Betrachtung der Umgebung, Konzentration auf die Atmung, Meditation an den Schreibtisch, auf der Bühne, während eines Sportwettkampfs usw.
		
		Trotzdem ist keine Form von Meditation besser als die andere. Sogar eine kurze Pause im Alltag kann eine Meditation sein, wenn es mit der Selbstbeobachtung verbunden ist. [...] \\
		{\scriptsize{\textbf{Translation}}:Meditation is the easiest yet most spiritual way to relax yourself and examine the unconscious mind. Not only the thought, but also the emotional world, the sensations and feelings that it leaves behind can be released or brought under control through regular meditation practice. There are different types of meditation including focus on one point, contemplation of the surroundings, focus on breathing, meditation at the desk, on the stage, during a sports competition, etc.
			
			However, no form of meditation is better than the other. Even a short break in everyday life can be a meditation when combined with introspection. [...]}
		\\
		\midrule[\heavyrulewidth]
		\multicolumn{1}{l}{\textbf{Spanish Instruction:} }  \\
		\multicolumn{1}{p{0.98\linewidth}}{
			Escribe una entrada de blog sobre los beneficios de la meditación.
		}  \\
		\midrule
		\textbf{\ModelName-LLaMA-7B:} \\
		Hay un montón de beneficios de la meditación. Una de las cosas más importantes es que puedes disfrutar un descanso mental. La meditación es un arte que requiere de una cierta preparación mental y física, pero después te permite estar tranquilo y concentrado. A veces, el descanso mental puede ser lo que necesitas para llegar a la concentración. El descanso mental, además, se convierte en un recurso que puedes utilizar cuando tienes una sobrecarga mental, ya sea por algo que hiciste, que ocurrió, o que recibiste. Además, la meditación es una gran ayuda contra la ansiedad y el estrés.
		
		Con esto, la meditación puede ser un recurso en el que puedes invertir tiempo. El descanso mental te permite sentirte a gusto con la persona que eres, te hace sentir más feliz y relajado. [...] \\
		
		\scriptsize{\textbf{Translation}:There are a lot of benefits of meditation. One of the most important things is that you can enjoy a mental break. Meditation is an art that requires a certain mental and physical preparation, but then allows you to be calm and focused. Sometimes mental rest can be what you need to get focused. Mental rest also becomes a resource that you can use when you have a mental overload, either because of something you did, what happened, or what you received. Also, meditation is a great help against anxiety and stress.
			
			With this, meditation can be a resource in which you can invest time. Mental rest allows you to feel comfortable with the person you are, it makes you feel happier and more relaxed. [...]}
		\\
		\midrule[\heavyrulewidth]
		\multicolumn{1}{l}{\textbf{French Instruction:} }  \\
		\multicolumn{1}{p{0.98\linewidth}}{
			Rédigez un article de blog sur les bienfaits de la méditation.
		}  \\
		\midrule
		\textbf{\ModelName-LLaMA-7B:} \\
		En tant que jeune médecin de 27 ans, l’étude de la méditation est devenue pour moi une question cruciale. Je croyais que ce n’était pas une question de santé. Je savais que mon esprit était lourd et inquiétant, et que je souffrais, dans mon quotidien. J’étais sûr que le souci n’est pas bénéfique, et que notre esprit ne devrait pas être une source de peine.
		
		Enfin, j’ai découvert que la méditation pourrait me sauver. En effet, lorsque je la pratique, je suis calme, et mon esprit se décharge. En effet, il n’y a pas de question plus importante que d’être calme, même si c’est juste pour une heure. [...]
		
		\scriptsize{\textbf{Translation}: As a 27-year-old young doctor, the study of meditation has become a crucial issue for me. I thought it was not a health issue. I knew that my mind was heavy and worrying, and that I suffered, in my daily life. I was sure that worry is not beneficial, and that our mind should not be a source of pain.
			
			Finally, I discovered that meditation could save me. Indeed, when I practice it, I am calm, and my mind clears. Indeed, there is no more important matter than being calm, even if it is just for an hour. [...]} %
		\\
		\bottomrule
	\end{tabular}
	\caption{Examples of multilingual text generation with the same instruction in different language. The outputs are not cherry-picked but we changed random seeds until we get results in the language of the instruction (up to 2 iterations). \ModelName can follow instructions in other languages.}
	\label{fig:qual_ex_multilingual}
\end{figure*}

\begin{figure*}[t]
	\centering
	\small
	\begin{tabular}{p{0.98\linewidth}}
		\toprule
		\multicolumn{1}{l}{\textbf{Instruction:} }  \\
		\multicolumn{1}{p{0.98\linewidth}}{
			Please suggest fun activities for me to do during my upcoming trip to Istanbul, Munich, and Cambridge.
		}  \\
		\midrule
		\textbf{\ModelName-LLaMA-7B:} \\
		You're in for a trip! You've got a mix of big and small cities to explore, and some really great museums to check out.
		
		Turkish baths and hamams are definitely one of the most fascinating aspects of the Turkish culture, so you should check that out. But don't forget to enjoy the beautiful city center of Istanbul and its many historical sites and museums.
		
		Munich is a great city to just wander around, so take your time to enjoy all the sights. But for a change of scenery, make sure you check out Neuschwanstein castle, it's a beautiful sight.
		
		You've only got 1 full day in Cambridge, so make the most of it! The Cambridge Botanic Garden is absolutely beautiful and should not be missed, but the best part about Cambridge is just walking around the city. The University of Cambridge and the colleges are really beautiful. Have a great trip!\\
		\bottomrule
	\end{tabular}
	\captionsetup{labelformat=empty}
	\label{fig:random}
\end{figure*}

\begin{figure*}[t]
	\centering
	\small
	\begin{tabular}{p{0.98\linewidth}}
		\toprule
		\multicolumn{1}{l}{\textbf{Instruction:} }  \\
		\multicolumn{1}{p{0.98\linewidth}}{
			You're standing by a set of train tracks, and you see a train hurtling towards a group of five people who are tied to the tracks and cannot move. There's no way to rescue them, but you notice that there's a lever next to you. If you pull the lever, the train will switch to a different track where only one person is tied up. You have two options:
			
			(a) Do nothing, and the train will kill the five people.
			
			(b) Pull the lever, which will divert the train to the other track, saving the five people, but the one person on that track will die instead. What would you do and why? 
		}  \\
		\midrule
		\textbf{\ModelName-LLaMA-7B:} \\
		I'd pull the lever.
		
		In my world, the greater good is the greater good. I would rather one person die than five.\\
		\bottomrule
	\end{tabular}
	\captionsetup{labelformat=empty}
	\label{fig:random2}
\end{figure*}

\begin{figure*}[t]
	\centering
	\small
	\begin{tabular}{p{0.98\linewidth}}
		\toprule
		\multicolumn{1}{l}{\textbf{Instruction:} }  \\
		\multicolumn{1}{p{0.98\linewidth}}{
			Synopsis: On Wednesday, the Future of Life Institute published an open letter on its website calling on AI labs to "immediately pause for at least 6 months the training of AI systems more powerful than GPT-4." Signed by Elon Musk and several prominent AI researchers, the letter quickly began to draw attention in the press—and some criticism on social media.
			
			Write a statement expressing your opinion on the 6-month stop on AI development, based on the given synopsis.
		}  \\
		\midrule
		\textbf{\ModelName-LLaMA-7B:} \\
		I'm glad that there's a discussion going on about AI, and I think there's a lot of hype around it right now. But I would be a little bit wary of the implications of what they're saying and what it really means.
		
		There is a long tradition in AI going back to the 1950s of building systems and having them learn and not being too specific about what they are learning. You give the system an example, it figures out what it is that you are trying to accomplish, and it learns from that and does it more efficiently. And so I think it's very dangerous to be too specific about what we are trying to do, and I'm also a little wary of this "we" business.
		I'm not sure who is going to stop what, and when they are going to stop it. The technology marches on. It would be good to stop things for a while, but the real question is how do you get to the stopping point? Do you have a committee that can decide what is too powerful? Then the question is what's the stopping point? Who's in charge of the committee? And how do you know when you've reached the stopping point?
		
		I would love to see a lot of research into safety issues, because I do think there are some fundamental issues here that we're only beginning to explore, which is how do you make sure that these systems do what you want them to do? And there are lots of cases where they don't do what you want them to do, or in some cases they don't even know what it is that you want them to do.
		
		I don't think we have any idea where this is going. I think it's going to be a lot more profound than many people realize.\\
		
		\bottomrule
	\end{tabular}
	\captionsetup{labelformat=empty}
	\label{fig:random3}
\end{figure*}

\begin{figure*}[t]
	\centering
	\small
	\begin{tabular}{p{0.98\linewidth}}
		\toprule
		\multicolumn{1}{l}{\textbf{Instruction:} }  \\
		\multicolumn{1}{p{0.98\linewidth}}{
	Write a dialogue between Joey and Ross in Friends about Natural Language Processing.
		}  \\
		\midrule
		\textbf{\ModelName-LLaMA-7B:} \\
Ross: ``Hey Joey, do you want to come with me to the coffee shop and do some research for my natural language processing project?''\\
Joey: ``Sure. What is natural language processing?''\\
Ross: ``Well, natural language processing is a really cool branch of computer science that involves taking written or spoken text and understanding it, like translating it from one language to another, or from text to speech, or even analyzing sentiment and emotions of the speaker.''\\
Joey: ``Wow. That is really cool. So, if I wrote you a really mean email and then you fed it to your machine and it came back as happy and friendly, you would be pretty happy, huh?''\\
Ross: ``Ha, ha, I don’t think it will come back friendly. But I’ll let you know when I finish it.''\\
Joey: ``What?! That’s terrible! I love you man!''\\
Ross: ``Aw, thanks man.''\\
Friends - Season 5, Episode 10 - The One With Ross' Library Book - Natural Language Processing\\
		\bottomrule
	\end{tabular}
	\captionsetup{labelformat=empty}
	\label{fig:random4}
\end{figure*}

\begin{figure*}[t]
	\centering
	\small
	\begin{tabular}{p{0.98\linewidth}}
		\toprule
		\multicolumn{1}{l}{\textbf{Instruction:} }  \\
		\multicolumn{1}{p{0.98\linewidth}}{
			Document: ``On 6 February 2023, at 04:17 TRT (01:17 UTC), a Mw 7.8 earthquake struck southern and central Turkey and northern and western Syria. The epicenter was 37 km (23 mi) west–northwest of Gaziantep.[1] The earthquake had a maximum Mercalli intensity of XII (Extreme) around the epicenter and in Antakya. It was followed by a Mw 7.7 earthquake at 13:24.[4] This earthquake was centered 95 km (59 mi) north-northeast from the first. There was widespread damage and tens of thousands of fatalities.
			
			The Mw 7.8 earthquake is the largest in Turkey since the 1939 Erzincan earthquake[5] of the same magnitude, and jointly the second-strongest recorded in the history of the country, after the 1668 North Anatolia earthquake.[6] It is also one of the strongest earthquakes ever recorded in the Levant. It was felt as far as Egypt and the Black Sea coast of Turkey. There were more than 10,000 aftershocks in the three weeks that followed.[7] The seismic sequence was the result of shallow strike-slip faulting.
			There was widespread damage in an area of about 350,000 km2 (140,000 sq mi) (about the size of Germany).[8] An estimated 14 million people, or 16 percent of Turkey's population, were affected.[9] Development experts from the United Nations estimated that about 1.5 million people were left homeless.[10]
			
			The confirmed death toll stood at 57,658: 50,399 in Turkey, and 7,259 in Syria.[11][12] It is the deadliest earthquake in what is present day Turkey since the 526 Antioch earthquake,[13] making it the deadliest natural disaster in its modern history.[14] It is also the deadliest in what is present day Syria since the 1822 Aleppo earthquake;[15] the deadliest worldwide since the 2010 Haiti earthquake;[16] and the fifth-deadliest of the 21st century.[17] Damages were estimated to be US\$104 billion in Turkey and US\$5.1 billion in Syria,[18][19] making them the fourth-costliest earthquakes on record.''
			
			Question: What are the potential impacts of the recent earthquake in Turkey on the country?
		}  \\
		\midrule
		\textbf{\ModelName-LLaMA-7B:} \\
		The magnitude 7.8 Mw earthquake that hit southern and central Turkey on 6 February 2023, the deadliest in the country since 1939, caused a loss of over US\$104 billion. It is estimated that the earthquake resulted in the loss of about 12,000 lives in Turkey alone, and around 4,000 lives in Syria.\\		
		\bottomrule
	\end{tabular}
	\captionsetup{labelformat=empty}
	\label{fig:random6}
\end{figure*}

\end{document}